\documentclass[runningheads]{llncs}
\usepackage{graphicx}
\usepackage{tikz}
\usepackage{comment}
\usepackage{amsmath,amssymb} 
\usepackage{color}
\usepackage{booktabs, multirow}
\usepackage{gensymb}
\usepackage{algorithmicx}
\usepackage{algcompatible}
\usepackage{algorithm}

\makeatletter
\newcommand{\printfnsymbol}[1]{%
  \textsuperscript{\@fnsymbol{#1}}%
}
\makeatother

\DeclareMathOperator*{\belowgeq}{\geq}

\begin{document}
\pagestyle{headings}
\mainmatter
\def\ECCVSubNumber{5671}  

\title{Learn distributed GAN with Temporary Discriminators} 

\titlerunning{Learn distributed GAN with Temporary Discriminators}
%
\author{Hui Qu\inst{1}\thanks{equal contribution} \and Yikai Zhang\inst{1}\printfnsymbol{1} \and Qi Chang\inst{1}\printfnsymbol{1} \and Zhennan Yan\inst{2} \and Chao Chen\inst{3} \and Dimitris Metaxas\inst{1}}
\authorrunning{H. Qu et al.}
%
\institute{Rutgers University, Piscataway, NJ 08854, USA \email{\{hq43,yz422,qc58,dnm}@cs.rutgers.edu\}
\and SenseTime Research \email{yanzhennan@sensetime.com} \and Stony Brook University, Stony Brook, NY 11794, USA \email{chao.chen.cchen@gmail.com}}

\maketitle
\begin{abstract}
In this work, we propose a method for training distributed GAN with sequential temporary discriminators. Our proposed method tackles the challenge of training GAN in the federated learning manner: How to update the generator with a flow of temporary discriminators? 
 We apply our proposed method to learn a self-adaptive generator with a series of local discriminators from multiple data centers. We show our design of loss function indeed learns the correct distribution with provable guarantees. The empirical experiments show that our approach is capable of generating synthetic data which is practical for real-world applications such as training a segmentation model. Our TDGAN Code is available at: https://github.com/huiqu18/TDGAN-PyTorch.
\end{abstract}

\section{Introduction}
\label{sec:intro}

\subsection{Advantages of distributed GAN learning}
In this work we focus on a practical framework for learning  Generative Adversarial Network (GAN)~\cite{goodfellow2014generative,arjovsky2017wasserstein,zhang2017stackgan,radford2015unsupervised} using \emph{multiple privately hosted discriminators from multiple entities}. Aggregating feedback from multiple local discriminators endues the generator a global perspective without accessing individual sensitive data. Such natural framework for distributed GAN learning is of great interest to both researchers and practitioners due to its advantages of privacy awareness and adaptivity.

\noindent\textbf{Privacy awareness} 
With more privacy concerns on sharing data and regulations imposed for privacy protection like HIPAA~\cite{annas2003hipaa,mercuri2004hipaa,milieu2014overview,gostin2009beyond} in medical domain, it is critical to consider privacy protection mechanism in designing a machine learning architecture~\cite{geyer2017differentially,konevcny2016federated}. 
The training framework of a distributed GAN using local discriminators meets the criterion of federated learning since the generator does not require access to sensitive data. Classical  federated learning framework~\cite{zhu2019deep,li2019federated,hard2018federated} shares gradients information, which is known to have information leakage. 
In distributed GAN framework, the local discriminator serves as a shield separating sensitive data from the querier thus is privacy comfortable. Besides, the nature of synthetic data allows the generator to share synthetic images without restriction, which is of great interest in privacy sensitive applications.

\noindent\textbf{Downstream task architecture adaptivity} 
Our proposed method could embrace the future upgrades of the downstream task by providing the well-trained image generator. The machine learning architecture upgrades rapidly to achieve a better performance by novel network modules~\cite{hoffman2016fcn,ronneberger2015u,milletari2016v,qu2019improving}, loss functions~\cite{sudre2017generalised,hochberg1964depth},  or optimizers~\cite{ruder2016overview,zeiler2012adadelta,mason2000boosting,zhang2019taming,zhang2020local}. However, the performance of the downstream task couldn't be better if any of the training datasets is missing. Our proposed method could compensate the risk of dataset missing by providing an image generator. The downstream task could trained on the synthetic images from generator without worrying about the loss of the proprietary datasets.

\subsection{Temporary datasets challenge for distributed GAN}

Several works have been done toward a practical framework to train generator by multiple  discriminators~\cite{hardy2019md,durugkar2016generative,chang2020synthetic}.  
However, all these methods assume discriminators/data centers are always available/online which is not realistic in the practical situation. In particular, those methods fail to consider the challenge of the temporary datasets problem. Since data centers are separated entities, privately hosted discriminators may join the collaborative learning in a stream fashion.  On one hand, new individual entities may join the network sequentially; more and more discriminators will participate. On the other hand, some local discriminators/nodes in the network may go offline and never come back again. For example, research funding for a hospital is limited, it is not realistic to require a hospital to maintain the dataset and stay online forever. Ideally, a general framework should be able to handle the temporary discriminator constraint, i.e. the dynamic flow of participating and leaving discriminators.

In the process of learning with flows of temporary discriminators, the dilution problem becomes a major concern. Suppose the generator keeps learning from late arrival discriminators, the memory of learned distribution with regard to absent data centers may be submerged by the incoming data. Such dilution phenomenon has been widely observed~\cite{mccloskey1989catastrophic}. A well designed framework should be able to aggressively capture the essence of online datasets while keeping the consistency of memory about the datasets of left parties.

To address such challenge in practice, we propose our training method called Temporary Discriminator GAN (TDGAN). Our method relies on a mixture of two loss functions: \emph{digesting loss} and \emph{reminding loss} to tackle the challenge. The digesting loss
updates the generator by collecting feedback from temporarily available and privately hosted discriminators. The reminding loss keeps the consistency of the generator on the distribution of the absent data centers thus prevents the memory decay issue of temporary discriminators. Our analysis shows the digesting loss and reminding loss can accomplish above tasks with provable guarantees. Hence minimizing the proposed joint loss allows the generator to approach the global distribution in a progressive fashion. 

To the best of our knowledge, this is the first work addressing the challenge of temporary discriminator problem in distributed GAN learning. Our main contributions in this paper include:
\begin{itemize}
    \item We propose a framework called TDGAN for training distributed GAN with temporary and privately hosted discriminators from multiple entities.
    \item Leveraging on two loss functions called digesting loss and reminding loss, our proposed framework enables the generator to learn a global distribution with temporary discriminators in a progressive fashion.
    \item We report an analysis on the digesting loss and reminding loss applied in TDGAN framework. Our analysis shows the proposed training framework leads the generator toward a correct distribution with provable guarantees.
\end{itemize}

 We empirically demonstrate the effectiveness of TDGAN on learning the global distribution, by applying TDGAN to pathology images generation and brain MRI images generation tasks. 

\section{Related Work}
\label{sec:related-work}
\subsection{Generative Adversarial Networks (GANs)}
The GANs proposed in \cite{goodfellow2014generative} seek to imitate data distribution via adversarial supervision from discriminator. Specifically, the generator focuses on generating  synthetic images indistinguishable to real images thus  the discriminator cannot effectively answer the `fake or real' queries. Such model and its variants have obtained great success in broad domains such as images~\cite{ledig2017photo,wang2018perceptual} , videos~\cite{tulyakov2018mocogan,vondrick2016generating}, music generation~\cite{guimaraes2017objective,lee2017seqgan}, natural languages~\cite{lin2017adversarial,hsu2017voice} and medical images~\cite{dai2018scan,mardani2017deep,yang2017automatic}.
In this work, we focus on conditional GAN~\cite{mirza2014conditiongan}, which aims to approximate the conditional distribution $p(x|y)$ given auxiliary variable $y$. In reality the $y$ is usually labels for classification data or masks for segmentation data. 

\subsection{Federated Learning}
Federated learning allows multiple data sources to train a model collaboratively without sharing the data~\cite{konevcny2016federated,hard2018federated}. Such setting protects the privacy of participants by communicating model information (e.g. model parameters, gradients) instead of original data~\cite{li2019federated}. In our setting, the federated learning framework is incorporated by separating local data and centralized generator using discriminators. In another word, only  discriminator kept locally has access to data and the only information communicated with cloud is the feedback on the synthetic data. In particular, \cite{hardy2019md} proposed a multiple-discriminator based  GAN for distributed training. However, their training framework requires swapping parameters between discriminators during the optimization process, which is not realistic in our sequential collaborative learning setting. 
In addition, nor these works provide a provable guarantee on the target distribution of generator given supervision from multiple  distributed discriminators. It is unclear whether their training framework can lead the generator toward the correct generative distribution. In this work, the proposed method addresses the problem of asynchrony problem meanwhile achieves provable guarantee on the target distribution. 

\subsection{Lifelong learning} 
Lifelong learning is the process of learning over time by accommodating new knowledge while remembering previously learned experiences, like how human learns. The main problem of lifelong learning for computational models like deep neural networks is the catastrophic forgetting~\cite{mccloskey1989catastrophic}, i.e., the model trained on new data is prone to perform bad in previous tasks. Many works have been proposed to solve this problem, including discriminative approaches~\cite{li2017learning,lopez2017gradient,aljundi2018memory} and generative models~\cite{seff2017continual,wu2018memory,zhai2019lifelong}. Our sequential collaborative learning setting is a type of lifelong learning since the forgetting problem also exists in our learning tasks. But it has two major differences from previous generative methods: 1) The discriminators of old tasks are unavailable when training with new data in our setting, and each discriminator only has access to its own dataset due to the privacy issue. 2) There may be multiple datasets/discriminators online for training at each time step and the number of discriminators is different from that of previous task. So it is not feasible to utilize the previous discriminators even if there is no privacy concern.

\section{Method}
\label{sec:method}

In this section we will first describe the problem definition and then present the details about the proposed TDGAN framework. We introduce how TDGAN solves the temporary datasets problem using the digesting loss and reminding loss. We also analyze the loss function of TDGAN and provide theoretical guarantees that it learns the correct distribution.

\subsection{Problem definition}
We first introduce the problem of learning distributed GAN with temporary discriminators. There are some local centers that host their own private datasets. The local centers/datasets are not always online. The target is to learn the mixture distributions of all local datasets without access to the real data. There is no assumption about the distribution of these local datasets, i.e., they can be \textit{i.i.d.} or \textit{non-i.i.d.} datasets. At each time step $t$, the task is to learn the distribution of current online datasets, and at the same time remember the learnt knowledge of previous offline datasets. Considering the privacy issue, each local dataset can only be accessed locally.

\subsection{TDGAN framework}
\begin{figure}[t]
	\begin{center}
		\includegraphics[width=11.5cm, height=9cm]{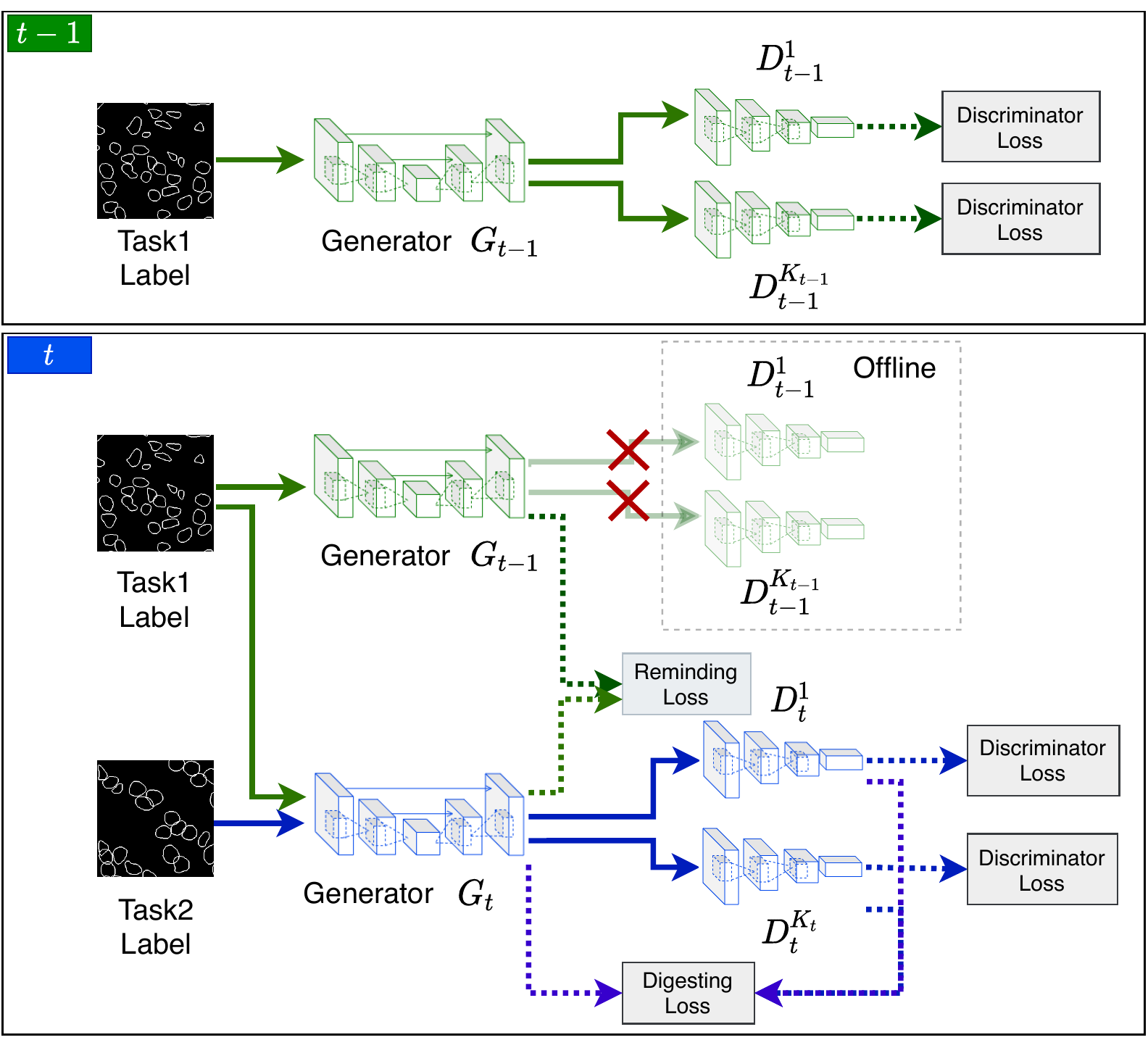}
	\end{center}
	\caption{The overview of TDGAN framework. 
	It consists of a central generator $G$ and multiple distributed temporary discriminators $\{D_{t}^1, \cdots, D_{t-1}^{K_{t-1}}, D_{t}^1, \cdots D_{t}^{K_{t}}, \cdots\}$. Each local data center has one discriminator. At time $t$, previous local discriminators are offline. $G_{t}$, initialized with $G_{t-1}$, tries to learn the data from newly online discriminators $\{D_{t}^1, \cdots, D_{t}^{K_{t}}\}$, and remember the knowledge learnt from previous data in $G_{t-1}$.}
	\label{arch}
\end{figure}

The overview of TDGAN framework is shown in Fig.~\ref{arch}. It contains a central generator and multiple distributed temporary discriminators located in different data centers (hospitals, mobile devices etc.). Each discriminator only has access to data stored in one local center, thus discriminators are trained in an asynchronous fashion.
Suppose the training starts at time $t-1$, there are $K_{t-1}$ online local data centers/discriminators. The central generator $G_{t-1}$ takes task-specific labels and random noise as inputs and outputs synthetic images to fool the discriminators. The local discriminators, $\{D_{t-1}^1, \dots, D_{t-1}^{K_{t-1}}\}$ learn to tell the synthetic images from the local real images. At time $t$, the real data and discriminators in local data centers of time $t-1$ are no longer available as new data comes in. The central generator $G_{t}$ tries to learn the distribution of new data and retain the mixture distribution learnt from previous data. The learning of new data is achieved by a digesting loss and the memory of previous learnt knowledge is kept using a reminding loss.

\noindent\textbf{Communication between $G$ and $D$s.} At each training step, the generator gets real labels sampled from online data centers, sends generated fake images to them, and then gets feedback from the discriminators in online data centers. For previous offline data centers, the generator has stored their real labels in the central server, and samples from the empirical distribution to generate fake images, which are used to remember previous learnt knowledge. More training and communication details can be found in the appendix.

\subsection{Loss function of TDGAN} 
The training framework is built based on the conditional GAN framework \cite{mirza2014conditiongan}. We aim to approximate the conditional distribution $p(x|y)$ given variable $y$. In reality, $y$ can be segmentation masks or classification labels. Given $y$, $x$ is the data generated from $p(x|y)$. We model the underlying mixture distribution of previous and current data at time $t$ as auxiliary variable $s_t(y)= (1-\alpha_t) s_{t-1}+ \alpha_t g_t(y) $.  Intuitively, new data source will join learning at each step $t$ thus the marginal distribution of auxiliary variable will be a mixture of existing and new distributions. In general, at each step $t$, $g_t(y)$ consists of multiple components $\{g^1_t,...,g^{K_t}_t\}$ with each component represents variables in separated data centers thus
$g_t(y)=\sum_{k=1}^{{K_t}} \pi^k_t g^k_t(y) $. The weight of each component can be computed via scaling the size of each data center, e.g., $\pi^k_t=\frac{n^k_t}{n_t}$ where $n_t=\sum_{k=1}^{{K_t}} n_t^k$. The generative distribution of $x$ given $y$ at time $t$ is $p_t(x,y)=  s_t(y) p(x|y)=\sum_{\tau=1}^{t} \alpha_{\tau}g_\tau(y)p(x|y)$. We assume the conditional distribution is consistent over time. In practice, we approximate $s_{t-1}(y)$ via the empirical distribution of $y$ sampled from data centers.  The loss function of TDGAN consists of two parts:
\begin{equation} \label{loss_main}
\begin{aligned}
V_t(G_t,D^{1:{K_t}}_t)=&\min\limits_{G_t} \;L_{Digesting}\;\; + \lambda  \cdot L_{Reminding}\;\; \;\;
\\
\text{\small{Digesting  Loss}}: L_{Digesting} \overset{\Delta}{=}&\;\max\limits_{D^{1:{K_t}}_t} \sum_{k=1}^{{K_t}} \pi^k_t \mathbb{E}_{y\sim g_t^k(y)} \left\{\mathbb{E}_{x\sim p(x|y)}[ \log D^k_t(x,y)] \right.\\
& + \left. \mathbb{E}_{u\sim unif(0,1)}[\log(1-D^k_t(G_{t}(u,y),y))]\right\}\\
\text{\small{Reminding  Loss}}: L_{Reminding} \overset{\Delta}{=}& \;\mathbb{E}_{y\sim s_{t-1}(y)}\mathbb{E}_{u\sim unif(0,1)}[ \|G_{t}(u,y)-G_{t-1}(u,y)\|^2]
\end{aligned}
\end{equation}
where $u$ is $unif (0,1)$ distribution and represents the noise input fed into the generator for synthetic data. The mixture cross entropy loss term provides a guidance for the generator to learn conditional distributions given auxiliary variable $y\in supp(g_t(y))$. In reality, $y\in supp(g_t(y))/supp(s_{t-1}(y))$ are the masks or labels that have not been observed before.  This is so called \emph{digesting loss}. The squared norm loss corresponds to the  \emph{reminding loss} which enforce the generator to memorize the conditional distribution of seen labels.

\subsection{Theoretical guarantees of TDGAN loss}
In this section we analyze the correctness of loss function. Due to the limit of space, all technical details are left in the supplementary materials. Ideally, the loss function should lead the generator toward the target distribution, formally $G_t(u,y)=p(x|y)$. To begin with, we first give some notations in the analysis. In order to describe the support incremental process in progressive collaborative learning, we use $\Omega_t$ be support of marginal distribution of auxiliary variables $\Omega(s_t(y))$ and $\Delta \Omega_t$ be the marginal increment of support, formally $\Delta \Omega(s_t(y))=\Omega(s_t(y))-\Omega(s_{t-1}(y))$. By above definition we have $\Delta \Omega_t= \Omega(g_t(y))-\Omega(s_{t-1}(y))$. Intuitively, the marginal incremental support contains previously unseen labels from new data center and we expect generator to mimic the distribution via interacting with discriminator locally trained in new data center. In addition, we call $\Omega(g_t(y))-\Omega(s_{t-1}(y))$ the absent support. The absent support contains auxiliary variables that are supervised by the discriminators in the offline data center in the past. In order to obtain supervision under the temporary discriminator constraint, the loss function is a mixture \emph{reminding loss} and \emph{digesting loss}. Our analysis of loss function consists two parts, which can be summarized as 1) digesting loss guides the generator to learn a correct  conditional distributions w.r.t auxiliary variables in $\Delta_t \Omega$. 2) reminding loss will enforce the generator to be consistent about the conditional distribution w.r.t auxiliary variables in the absent support. In sum, by the analysis in this section we aim to show that the digesting loss and reminding loss interacts in a learn and review manner. Even without feedback from off-line data centers, the generator can still manage to memorize learned distribution. We will begin with two lemmas.

\begin{lemma} [\textbf{Reminding loss enforces consistency}] \label{optrmd}
	Suppose $G_t$ has enough model capacity, the optimal $G_t$ for loss function:\\
	\begin{equation*}
	\min\limits_{G_t}\mathbb{E}_{y\sim s_{t-1}(y)}\mathbb{E}_{u\sim unif(0,1)}[ \|G_{t}(u,y)-G_{t-1}(u,y)\|^2]
	\end{equation*}
	given $G_{t-1}$ is  $G_t(u,y)=G_{t-1}$ for all $u$ and $y \in \Omega(s_{t-1}(y))$.
\end{lemma}

\noindent
\begin{lemma} [\textbf{Digesting loss learns correct distribution}]\label{optgan}
	Suppose discriminator $D^{k}_t, k\in [{K_t}]$ always behave optimally and let $q_t(x|y)$ be the distribution of $G_t(u,y)$, the  the optimal $G_t(u,y)$ for digesting loss:
	\begin{equation*}
	\begin{aligned}
	\min\limits_{G_t}&\max\limits_{D^{1:{K_t}}_t} \sum_{k=1}^{{K_t}} \pi^k_t\mathbb{E}_{y\sim g_t^k(y)} \{\mathbb{E}_{x\sim p(x|y)}[ \log D^k_t(x,y)]\\
	&+ \mathbb{E}_{u\sim unif(0,1)}[\log(1-D^k_t(G_{t}(u,y),y))]\}
	\end{aligned}
	\end{equation*}
	is $q_t(x|y)=p(x|y)$ for all $y \in \Omega(g_t(y))$.
\end{lemma}

The two lemmas describe the behavior of digesting loss and reminding loss separately. In next theorem, we show that the design of loss can work cooperatively thus the overall loss function leads to a correct global distribution.
\begin{theorem}
	Suppose the generator has enough model capacity to obtain $q_1(x|y)=p(x|y)$ for all  $y \in g_1(y)$ and the loss $V_\tau(G_\tau,D_\tau)$ defined in Equation \ref{loss_main} is optimized optimally for each $\tau \in [t]$, then $q_t(x|y) =p(x|y)$ for all $y\in \Omega_t$.
\end{theorem}

\noindent\textbf{Proof sketch}:\\
The proof is based on Lemma \ref{optrmd} and \ref{optgan} and the fact that optimal condition of reminding loss and digesting loss doesn't contradict each other for $y \in \Omega(s_{t-1}) \cap \omega(g_t)$. Using induction we show if each step TDGAN stops with optimal condition at step $\tau \in \{1,...,t-1\}$, eventually $q_t(x|y)=p(x|y)$ at time $t$. \qed

\begin{remark}
	The analysis shows the digesting loss and reminding loss control the behavior of generator w.r.t auxiliary variables in different regimes. The analysis relies on the fact that for $y$ in all regimes, the conditional distribution $p(x|y)$ is consistent over time.  Such assumption avoids the conflict conditional distribution case i.e. $p_\tau(x|y) \neq p_{\tau'}(x|y)$. We believe this assumption is necessary for success of learning. Otherwise conflict conditional distribution may confuse the generator.
\end{remark}

\section{Experiments}
\label{sec:exp}

In this section, we evaluate the power of TDGAN framework in distributed GAN learning problems. We focus on medical datasets in health entities which are known to be privacy sensitive. 
We perform experiments on pathology image generation and brain MRI image generation tasks to illustrate that TDGAN can learn the new distribution while keeping consistency of memory on learnt distributions. Since our generator is used as a synthetic database and can provide images for downstream tasks, we evaluate the quality of generated data using segmentation results of models trained by synthetic data. The TDGAN is compared with rule of thumb methods, e.g., fine-tuning and joint learning. The fine-tuning serves as a lower bound for TDGAN since it is a naive way in the sequential learning setting. Fine-tuning can also show the importance of the reminding loss because it only uses the digesting loss for GAN training. The joint learning method directly aggregates data from multiple entities and train a GAN in a centralized manner. Compared to the setting of TDGAN, such setting has less constraint thus serve as a upper bound of our result when the datasets of different entities are homogeneous. In the cases where the datasets varies a lot, our TDGAN can better learn the mixture distribution with the assists of distributed discriminators than joint learning.


\subsection{Experimental Set-up}
\subsubsection{Datasets}
Two segmentation datasets are used in the experiments.
\paragraph{Multi-Organ (MO)}
This is a nuclei segmentation dataset proposed in~\cite{kumar2017dataset}. It consists of 30 pathology images from seven organs. The training set contains 16 images from liver, breast, kidney and prostate. The same organ test set has 8 images from the above four organs and the different organ test set contains 6 images from bladder, colon and stomach. 


\paragraph{BraTS}
It is the dataset of the Multimodal Brain Tumor Segmentation Challenge 2018~\cite{bakas2017advancing,bakas2018identifying,menze2014multimodal}. Each patient in the dataset has four types of MRI scans: 1) a
native T1-weighted scan (T1), 2) a post-contrast T1-weighted scan (T1Gd), 3) a native T2-weighted scan (T2), and 4) a T2 Fluid Attenuated Inversion Recovery (T2-FLAIR) scan~\cite{bakas2018identifying}. And the annotation contains three types of tumor subregion labels: 1) active tumor (AT), 2) tumor core (TC), and 3) whole tumor (WT). In our experiments, we generate two datasets (BraTS-T1, BraTS-T2) from this dataset that differ much in the training images. To avoid large imbalance between datasets in different local centers, we select the T1 scans of 17 high-grade glioma (HGG) cases as the training data of BraTS-T1, and select the T2 scans of another 17 HGG cases as the training data of BraTS-T2. The test sets in both datasets contain 40 cases, and share the same labels but differ in the images (T1 vs. T2). For the annotation, we only use the whole tumor in both GAN training and segmentation for simplicity.

\subsubsection{Two types of tasks}
In the setting of temporary discriminators, the previous discriminators will be unavailable when new hospitals/datasets are online (see Fig.~\ref{arch}). We divide the image generation tasks into two types according to the difference between the datasets of hospitals.

\paragraph{Homogeneous tasks}
In this type of tasks, images in all hospitals' datasets are homogeneous, i.e., they have similar types and appearances. In our experiments, we assume the data in each local center comes from the Multi-Organ dataset.

\paragraph{Heterogeneous tasks}
In these tasks, images of different hospitals are heterogeneous, e.g., CT and MRI. To better illustrate the effects of our framework, we assume that the whole Multi-Organ dataset is in one hospital, the BraTS-T2 dataset in the second hospital and the BraTS-T1 dataset is in the third one. The distributions of these datasets are diverse, thus it is harder for the network to remember previous tasks and achieving good results on the new task.

\subsubsection{Evaluation metrics}
For nuclei segmentation, we use Dice score to measures the overlap between ground-truth mask $G$ and segmented result $S$: $Dice(G, S) = \frac{2|G \cap S|}{|G| + |S|}$.
Because nuclei segmentation is an instance segmentation problem, we use an additional object-level metric Aggregated Jaccard Index (AJI)~\cite{kumar2017dataset}:
\begin{equation}
AJI = \frac{\sum_{i=1}^{n_\mathcal{G}} |G_i \cap S(G_i)|}{\sum_{i=1}^{n_\mathcal{G}}|G_i \cup S(G_i)| + \sum_{k\in K}|S_k|}
\end{equation}
where $S(G_i)$ is the segmented object that has maximum overlap with $G_i$ with regard to Jaccard index, $K$ is the set containing segmentation objects that have not been assigned to any ground-truth object.

For brain tumor segmentation, we use Dice score and the 95\% quantile of Hausdorff distance (HD95). The Hausdorff distance is defined as
\begin{equation}
HD(G, S) = \max\{\sup_{x\in\partial G}\inf_{y\in\partial S}d(x, y), \sup_{y\in\partial S}\inf_{x\in\partial G}d(x, y)\}
\end{equation}
where $\partial$ means the boundary operation, and $d$ is Euclidean distance. The Hausdorff distance is sensitive to small outlying subregions, therefore we use the 95\% quantile of the distances as in~\cite{bakas2018identifying}.

\subsubsection{Implementation details}
\paragraph{Network structure}
In the GAN training phase, the central generator is an encoder-decoder network that consists of two stride-2 convolutions (for down-sampling), nine residual blocks \cite{he2016resnet}, and two transposed convolutions. All non-residual convolutional layers are followed by batch normalization \cite{ioffe2015batch} and the ReLU activation \cite{agarap2018deep}. All convolutional layers use $3\times3$ kernels except the first and last layers that use $7\times7$ kernels. Each discriminator has the same structure as that in PatchGAN~\cite{isola2017image} with patch size 70$\times$70. The segmentation model is U-net~\cite{ronneberger2015u}.

\paragraph{Data augmentation}
In the GAN training phase, we resize the image to 286$\times$286 and randomly crop to 256$\times$256 for training. In the nuclei segmentation tasks, each large 1000$\times$1000 image is split into 16 small 256$\times$256 patches. The other augmentations are random cropping of size 224$\times$224, random horizontal flip, randomly resize between 0.8 and 1.25, rotation within a random degree between -90$\degree$ and 90$\degree$, perturbation of the affine transform's parameters with a random value between -0.3 and 0.3. For brain tumor segmentation, the operations include random cropping of size 224$\times$224, random horizontal flip with probability 0.5 and rotation within a random degree between -10$\degree$ and 10$\degree$.

\begin{table}[t]
	\caption{Nuclei segmentation results using the models trained on generated images in single-entity nuclei image synthesis.}
	\label{tab:nuclei1}
	\begin{center}
		\begin{tabular}{l|cc|cc|cc|cc}
			\hline
			\multirow{2}{*}{Tasks (online dataset)} & \multicolumn{2}{c|}{TDGAN}  & \multicolumn{2}{c|}{Fine-tuning}  & \multicolumn{2}{c|}{Joint learning} & \multicolumn{2}{c}{Local GAN}  \\ \cline{2-9}
			   & Dice $\uparrow$ & AJI $\uparrow$  & Dice $\uparrow$ & AJI $\uparrow$  & Dice $\uparrow$ & AJI $\uparrow$ & Dice $\uparrow$ & AJI $\uparrow$  \\	\hline
			Task1 (liver) & 0.6676  & 0.3420 & 0.6676 & 0.3420 & 0.6676 & 0.3420 & 0.6676 & 0.3420\\
			Task2 (breast) & 0.6961  & 0.4323 & 0.6950 & 0.4405  & 0.7114 & 0.4457 & 0.6745 & 0.4111 \\
			Task3 (kidney) & 0.7164 & 0.4512 & 0.7142 & 0.4195 & 0.7350 & 0.4814 & 0.6794 & 0.3734\\
			Task4 (prostate) & 0.7481 & 0.4931 & 0.6969 & 0.4679 & 0.7627 & 0.5184 & 0.6918 & 0.4401\\
			\hline
		\end{tabular}
	\end{center}
\end{table}

\begin{table}[t]
	\caption{Nuclei segmentation results using the models trained on generated images in multiple-entity nuclei image synthesis.}
	\label{tab:nuclei2}
	\begin{center}
		\begin{tabular}{l|cc|cc|cc|cc}
			\hline
			\multirow{2}{*}{Tasks (online dataset)} & \multicolumn{2}{c|}{TDGAN}  & \multicolumn{2}{c|}{Fine-tuning}  & \multicolumn{2}{c|}{Joint learning}  & \multicolumn{2}{c}{Local GAN} \\ \cline{2-9}
			   & Dice $\uparrow$ & AJI $\uparrow$  & Dice $\uparrow$ & AJI $\uparrow$  & Dice $\uparrow$ & AJI $\uparrow$  & Dice $\uparrow$ & AJI $\uparrow$  \\	\hline
			Task1 (liver+breast) & 0.6829  & 0.4291 & 0.6829 & 0.4291 & 0.7114 & 0.4457 & 0.6298 & 0.3813 \\
			Task2 (kidney+prostate) & 0.7599  & 0.5136 & 0.7285 & 0.4780  & 0.7627 & 0.5184 & 0.7187 & 0.4711\\
			\hline
		\end{tabular}
	\end{center}
\end{table}

\paragraph{Training details}
In the GAN training, we use the Adam optimizer~\cite{kingma2014adam} with a batch size of 8, and momentum parameters $\beta_1$ = 0.5, $\beta_2$ = 0.999. The number of epoch is set to 300 in homogeneous tasks and 80 in heterogeneous tasks. The learning rate is 0.0002 for the first half number of epochs and then linearly decayed. The parameters in the loss function (Eqn.~(\ref{loss_main})) is set to $\lambda=1$. In the segmentation phase, we train the U-net with Adam optimizer with a batch size of 12, using a learning rate of 0.001 for 50 epochs in brain tumor segmentation and 100 epochs in nuclei segmentation.

\subsection{Results on Homogeneous Tasks}\label{exp_homo}
In this subsection we show that our TDGAN can learn the overall distribution well when the data from different health entities are homogeneous.

\subsubsection{Settings}
The training data of Multi-Organ dataset is divided into four subsets according to organs: liver, breast, kidney and prostate. Each subset is assumed to be in a local health entity, consisting of 64 images of size 256$\times$256. We perform the following experiments: (1) \textbf{TDGAN}. The health entities are temporarily available, we train a generator using the proposed TDGAN method.  (2) \textbf{Sequential fine-tuning}. The health entities are temporarily available, the generator is fine-tuned in a sequential manner. Because previous discriminators are offline, only the generator is initialized using parameters from the generator trained/fine-tuned on the previous tasks. The new local discriminators are randomly initialized. (3) \textbf{Joint learning}. The data in each health center can be collected together to train a regular GAN model. (4) \textbf{Local GAN}. A local GAN is trained using the local data for each health entity.

To evaluate the performance of generators, synthetic images are generated using labels from both previous data and current data. Then they are used to train a segmentation model. The higher accuracy the trained segmentation model performs on the test set, the better quality of the synthetic images have. To make fair comparison, the number of synthetic images keeps the same for each segmentation model. We only use the same organ test set for evaluation.

Besides, we set the number of online health entities at each time to 1 and 2, corresponding to single-entity and multiple-entity cases, respectively. There are four tasks in the former case and two tasks in the latter one.

\begin{figure}[t]
    \centering
    \includegraphics[width=\textwidth]{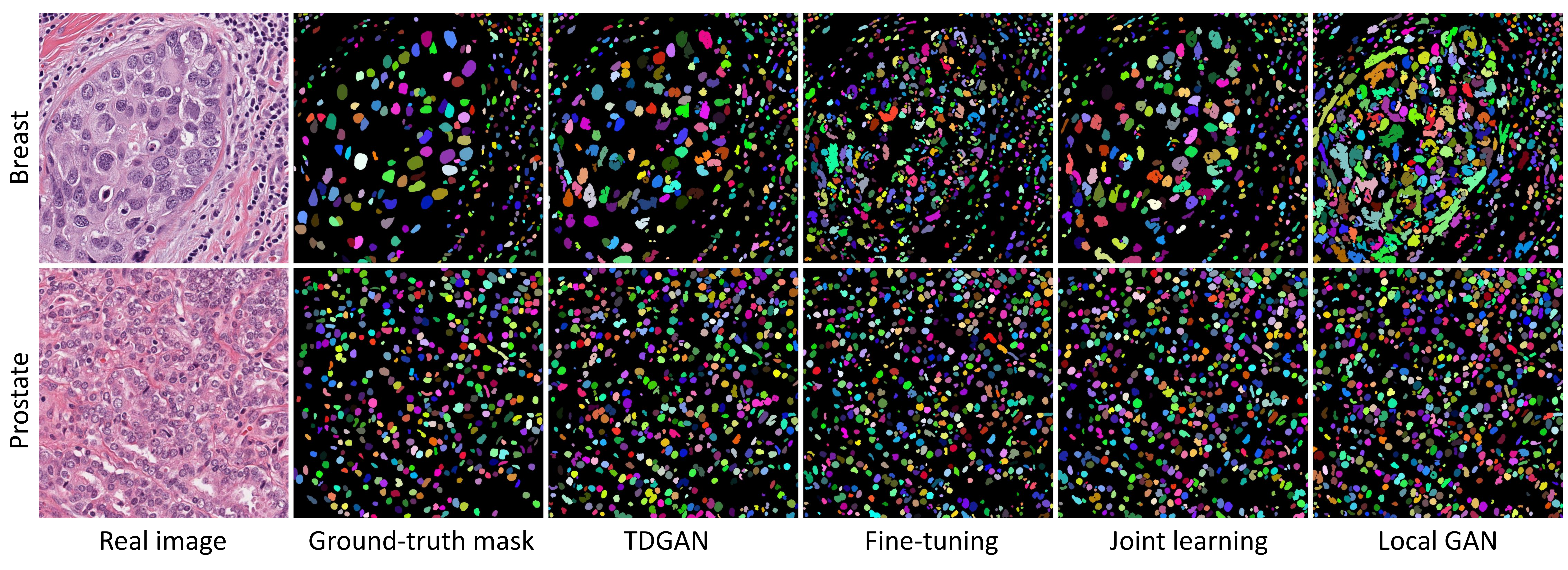}
    \caption{Segmentation results of models trained on synthetic images in task4 of single-entity nuclei synthesis. The first row is a breast image in test set, corresponding to the dataset of task2. The second row is a prostate image corresponding to the dataset of task4. TDGAN learns the mixture distribution, thus performs well on both images.}
    \label{fig:nuclei-vis}
\end{figure}

\begin{figure}[t]
    \centering
    \includegraphics[width=\textwidth]{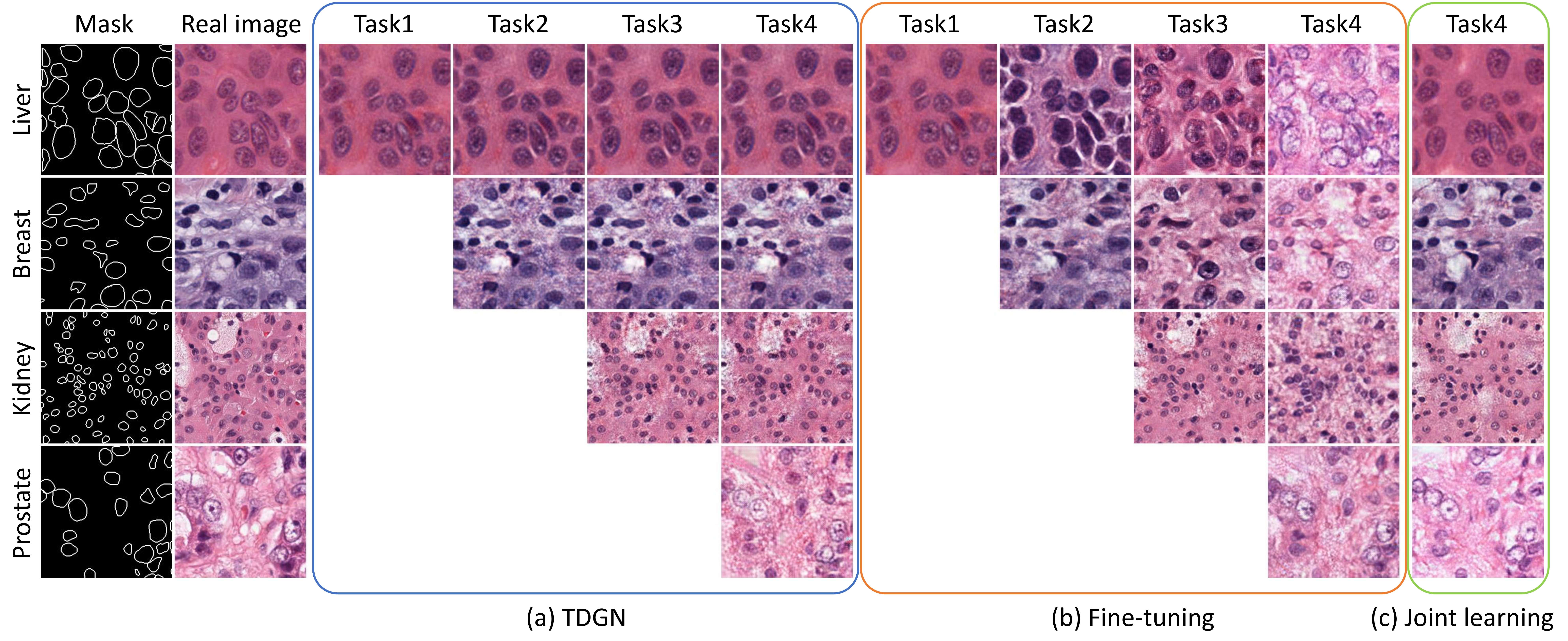}
    \caption{Comparison between different methods for single-entity homogeneous tasks. All methods share the model trained on the first dataset (Liver). During each subsequent task, the fine-tuning method forgets previous task(s), while our TDGAN can remember previous task(s) and learn the current task well.}
    \label{fig:nuclei-cmp}
\end{figure}
\begin{figure}[!h]
    \centering
    \includegraphics[width=0.7\textwidth]{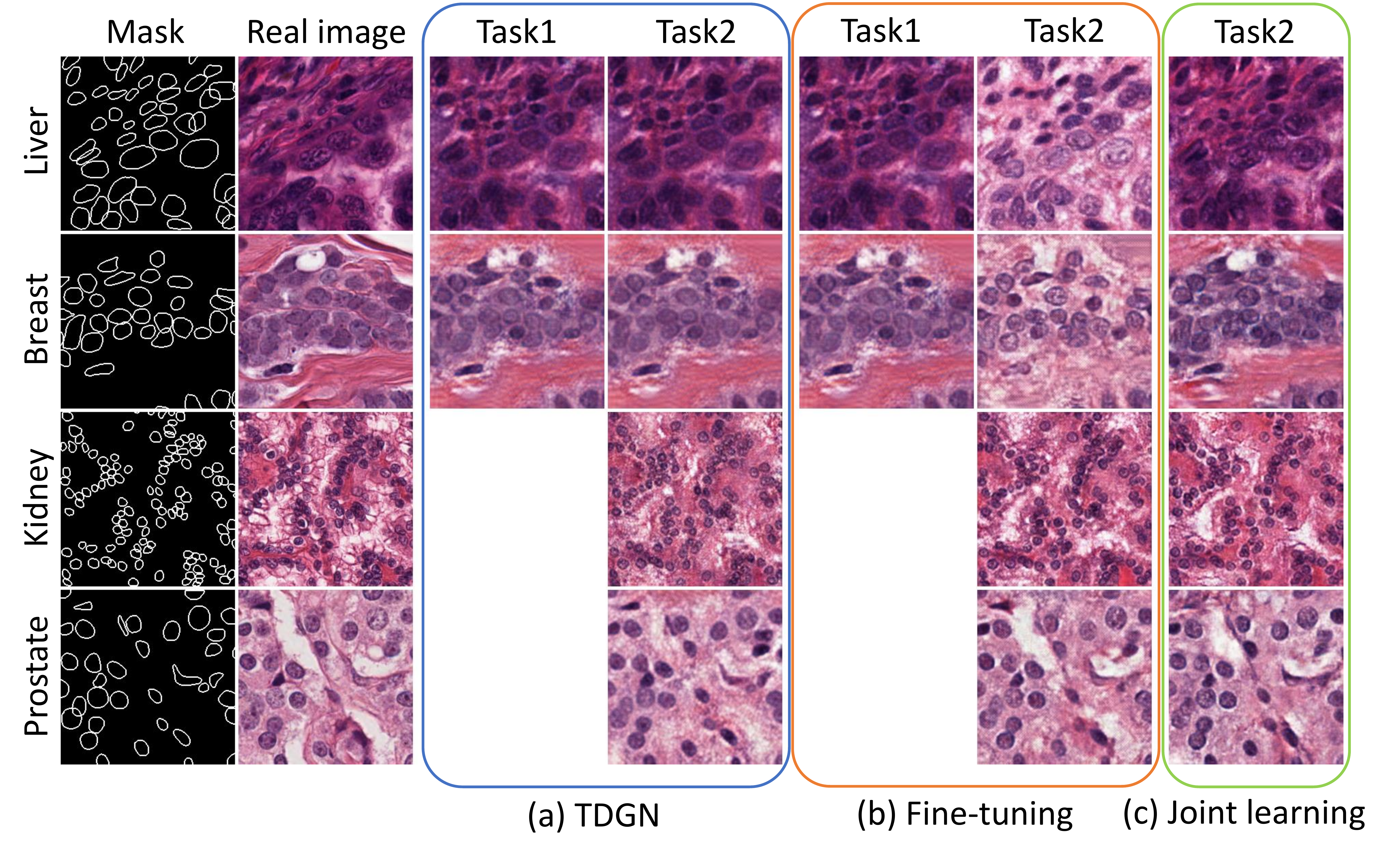}
    \caption{Comparison between different methods for multiple-entity homogeneous tasks. TDGAN and fine-tuning share the distributed GAN model trained on the first task (Liver+Breast), but joint learning train a regular GAN. After trained on the second task (Kidney+Prostate), the fine-tuning method forgets the first task, while our TDGAN can remember previous task and learn the current task well.}
    \label{fig:nuclei-cmp2}
\end{figure}

\subsubsection{Results}

The segmentation results trained with synthetic images under different settings are presented in Table~\ref{tab:nuclei1}, Table~\ref{tab:nuclei2} and Fig.~\ref{fig:nuclei-vis}. Some typical synthetic images are shown in Fig.~\ref{fig:nuclei-cmp} and Fig.~\ref{fig:nuclei-cmp2}. Our TDGAN outperforms fine-tuning on both single-entity and multiple-entity cases. Because the fine-tuning method focuses on learning current data using the digesting loss only, which affects the quality of synthetic images on previous datasets (Fig.~\ref{fig:nuclei-cmp}(b) and Fig.~\ref{fig:nuclei-cmp2}(b)). The proposed TDGAN not only learns the distribution of new data using the digesting loss, but also remembers the old data due to the reminding loss. TDGAN also outperforms Local GAN method because it can make use of the distributed datasets, which is the advantage of federated learning. Joint learning is the best among all methods. It is the upper bound of TDGAN in the homogeneous tasks.

Another observation is that the TDGAN and joint learning methods obtain better segmentation results as new data comes online, since more data is beneficial for GAN training. However, it is expected that the increase speed will slow down and the performance may even oscillate for TDGAN after a long time. Because TDGAN is still forgetting, and the errors will accumulate as time goes on. For fine-tuning, the performance has begun to oscillate at task 4 and will drop as it forgets more and faster.

\subsection{Results on Heterogeneous Tasks}
In this subsection we show that our TDGAN can learn the different distributions when the data from health entities are heterogeneous.
\begin{table}[t]
	\caption{Segmentation results using the models trained on generated images in heterogeneous image synthesis. The training datasets for Task1$\sim$3 are Multi-Organ, BraTS-T2 and BraTS-T1, respectively.}	\label{tab:brain}
	\begin{center}
		\begin{tabular}{c|l|cc|cc|cc|cc}
			\hline
		 \multirow{2}{*}{Test set} & \multirow{2}{*}{Tasks} &  \multicolumn{2}{c|}{TDGAN}  & \multicolumn{2}{c|}{Fine-tuning}  & \multicolumn{2}{c|}{Joint learning} & \multicolumn{2}{c}{Local GAN} \\ \cline{3-10}
			   &   & Dice $\uparrow$ & AJI $\uparrow$  & Dice $\uparrow$ & AJI $\uparrow$  & Dice $\uparrow$ & AJI $\uparrow$  & Dice $\uparrow$ & AJI $\uparrow$  \\	\hline
		\multirow{3}{*}{Multi-Organ} & Task1 & 0.7576 & 0.5151  & 0.7576 & 0.5151 &0.7576 & 0.5151 & 0.7576 & 0.5151  \\
			& Task2  & 0.7570 & 0.5180 & 0.3733 & 0.0980 & 0.6799 & 0.4029 & - & - \\
			& Task3  & 0.7610 & 0.5028 & 0.1812 & 0.0240 & 0.6566 & 0.4002 & - & - \\
			\hline\hline
			 &  & Dice $\uparrow$ & HD95 $\downarrow$   & Dice $\uparrow$ & HD95 $\downarrow$   & Dice $\uparrow$ & HD95 $\downarrow$   & Dice $\uparrow$ & HD95 $\downarrow$    \\	\hline
		\multirow{2}{*}{BraTS-T2} &  Task2 & 0.6834 & 37.23  & 0.6667  & 33.74 & 0.6734 & 40.33 & 0.5734 & 63.33  \\
			 & Task3  & 0.6713 & 36.33  & 0.2490 & 74.13 & 0.7027 & 33.54  & - & - \\ \hline\hline
			 BraTS-T1 & Task3 & 0.5265 & 36.98 & 0.5288 & 38.86 & 0.4627 & 53.86 & 0.4604 & 47.68 \\ \hline
		\end{tabular}
	\end{center}
\end{table}

\subsubsection{Settings}
In this case we have three health entities and each has one of Multi-Organ, BraTS-T2 and BraTS-T1 datasets. To make the model structure consistent, we adopt 2D images synthesis and segmentation for BraTS-T1 and BraTS-T2 datasets although they are 3D data. We extract 2D slices from each 3D volume for both training and test sets. To avoid severe imbalance problem during training, there are only 17 cases are selected in both BraTS datasets. Finally, there are 256, 1052 and 1110 images in the the three datsets, respectively.

We conduct similar experiments as in Section~\ref{exp_homo}. The difference in GAN training is that we didn't perform multiple-entity experiments, because of the large variance among three datasets. For segmentation tasks, we evaluate the synthetic images on the corresponding test set instead of using a global test set.

\begin{figure}[t]
    \centering
    \includegraphics[width=\textwidth]{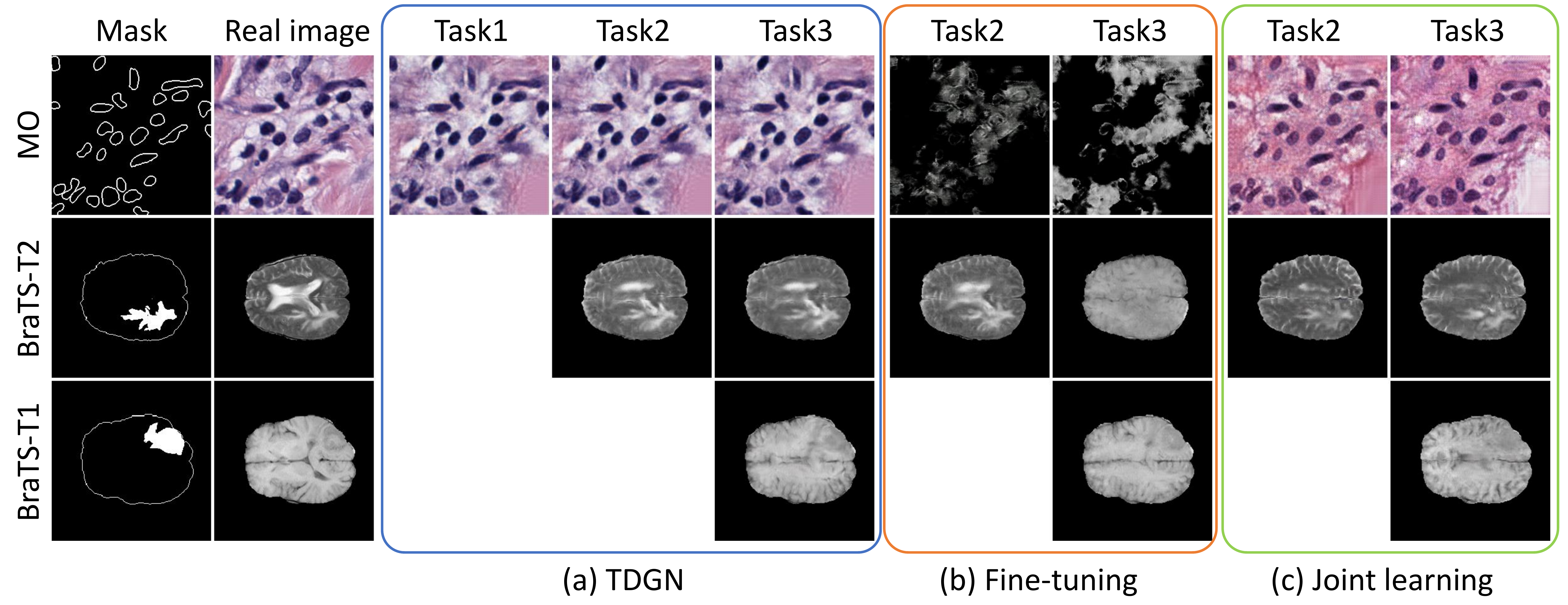}
    \caption{Comparison between different methods for heterogeneous tasks.  All methods share the model trained on the first dataset (Multi-Organ). During each subsequent task, the fine-tuning method forgets previous task(s), joint learning cannot deal with the large variance between datasets very well. Our TDGAN can handle both issues and learn the mixture distribution.}
    \label{fig:brain-cmp}
\end{figure}

\subsubsection{Results}
The segmentation results trained with synthetic images under different settings are presented in Table~\ref{tab:brain}. An example of synthetic images during the training process is shown in Fig.~\ref{fig:brain-cmp}. After trained with new heterogeneous data (Task2 and Task3), TDGAN's performance almost keeps the same on the test sets, while fine-tuning drops a lot on previous tasks, indicating that TDGAN can better preserve the memory of previous tasks. The performance of joint learning on MO dataset also deteriorates, because the large variance in heterogeneous datasets makes it hard to learning the mixture distribution using only one discriminator. In our TDGAN framework, each local discriminator is responsible for its own data, which is beneficial for learning in this case. TDGAN is also better than local GAN, because of more training data from different centers.

\section{Conclusion}
\label{sec:conclude}
In this work, we proposed a framework for training
distributed GAN called TDGAN. Our proposed training method allows the generator to learn from  temporarily available and privately hosted discriminators from multiple data centers.
TDGAN is a leverage on two loss functions called digesting loss and reminding loss to balance between learning new distribution and memorizing learned distributions. We evaluate the quality of the generator via the accuracy of segmentation model trained solely by synthetic data.  
Our empirical results show TDGAN achieves better performance than the model fine-tuned on temporary datasets, and achieves comparable performance as the model learns from joint  real homogeneous image datasets. When there is large variance in the datasets, our method can better learn the mixture distribution than joint learning.

\paragraph{\textbf{Acknowledgement}}
We thank anonymous reviewers for helpful comments. The research of Chao Chen is partially supported by NSF IIS-1855759, CCF-1855760 and IIS-1909038. The research of Dimitris Metaxas is partially supported by NSF CCF-1733843, IIS-1763523, CNS-1747778, and IIS-1703883.

\bibliographystyle{splncs04}
\bibliography{egbib}

\newpage
\section*{Appendix}
\renewcommand{\thesubsection}{\Alph{subsection}}

\subsection{Training algorithm of TDGAN}
The training algorithm and communication details between the central generator and distributed discriminators of TDGAN are shown in Algorithm 1.

\begin{algorithm}[!ht] 	
	\label{algorithm1}
	\caption{\small Training algorithm of TDGAN at time step $t$.}
	\begin{algorithmic}[1]
		\STATE{Initialized $G_t$ with $G_{t-1}$ if $t>1$.}
		\FOR{number of total training iterations}
		\STATEx{ \quad// \textit{Update online Discriminators}}
		\FOR{each online node $k \in [K_t]$} 
		\STATE{-- Sample  minibatch of of $m$ variables $\{y^k_1,...,y^k_m\}$ from $g^k_t(y)$.}
		\STATE{-- Send the minibatch from $D_t^k$ to $G_t$.}
		\STATE{-- Generate $m$ fake data from $G_t$, $\{\hat{x}^k_1,...,\hat{x}^k_m\}\sim q_t(\hat{x}|y)$.}
		\STATE{-- Send the fake data from $G_t$ to $D_t^k$.}
		\STATE{-- Update the discriminator $D_t^k$ by ascending its stochastic gradient:
			\vspace{-0.5em}
			\[\nabla_{\theta_{D_t^k}} \frac{1}{m} \sum_{i=1}^m \left[
			\log D_t^k(x_i^k)
			+ \log (1-D_t^k(\hat{x}_i^k))
			\right].
			\]}
		\vspace{-1.5em}
		\ENDFOR
		\STATEx{\quad// \textit{Compute the gradients of $G_t$ using the digesting loss}}
		\FOR{each online node $k \in [K_t]$}
		\STATE{-- Sample  minibatch of $m$ variables $\{y^k_1,...,y^k_m\}$ from $g_t^k(y)$.}
		\STATE{-- Send the minibatch from $D_t^k$ to $G_t$.}
		\STATE{-- Generate $m$ fake data from $G_t$, $\{\hat{x}^k_1,...,\hat{x}^k_m\}\sim q_t(\hat{x}|y)$.}
		\STATE{-- Send the fake data from $G_t$ to $D_t^k$.}
		\STATE{-- Collect error from $D_t^k$ for $G_t$.}
		\ENDFOR
		\STATE{-- Compute gradients on the digesting loss:
			\vspace{-0.5em}
			\[	\nabla_{\theta_{G_t}} \frac{1}{m} \sum_{k=1}^{K_t} \pi^k_t \sum_{i=1}^m
			\log (1-D_t^k(\hat{x}^k_i)).\]}
		\STATEx{\quad// \textit{Compute the gradients using the reminding loss if $t>1$}}
		\IF{$t>1$}
		\STATE{-- Sample  minibatch of $n$ variables $\{y_1,...,y_n\}$ from $s_{t-1}(y)$. (We approximate $s_{t-1}(y)$ by storing the empirical distribution in central server)}
		\STATE{-- Generate $n$ copies of $u$ from $unif(0,1)$: $\{u_1,...,u_n\}$.} 
		\STATE{-- Compute gradients on the reminding loss:
			\vspace{-0.5em}
			\[\nabla_{\theta_{G_t}} \frac{1}{n} \sum_{i=1}^n \|G_t(u_i,y_i) -G_{t-1}(u_i,y_i)\|^2.\]}
		\vspace{-1em}
		\ENDIF
		\STATE{Update $G_t$ using gradients from both losses.}
		\ENDFOR
	\end{algorithmic}
	\vspace{-0.3em}
\end{algorithm}

\subsection{Loss function of TDGAN}

\begin{equation} \label{loss_main}
\begin{aligned}
V_t(G_t,D^{1:{K_t}}_t)=&\min\limits_{G_t} \;Digesting\;\; Loss+ \lambda  \cdot Reminding\;\; Loss\;\;
\\
Digesting\;\; Loss \overset{\Delta}{=}&\;\max\limits_{D^{1:{K_t}}_t} \sum_{k=1}^{{K_t}} \pi^k_t             
\mathbb{E}_{y\sim g_t^k(y)} \{\mathbb{E}_{x\sim p(x|y)}[ \log D^k_t(x,y)]\\
& +\mathbb{E}_{u\sim unif(0,1)}[\log(1-D^k_t(G_{t}(u,y),y))]\}\\
Reminding\;\; Loss \overset{\Delta}{=}& \;\mathbb{E}_{y\sim s_{t-1}(y)}\mathbb{E}_{u\sim unif(0,1)}[ \|G_{t}(u,y)-G_{t-1}(u,y)\|^2]
\end{aligned}
\end{equation}

\subsection{Missing Proof in Analysis Section}

\begin{lemma} [\textbf{Reminding Loss enforces consistency}] \label{optrmd}
	Suppose $G_t$ has enough model capacity, the optimal $G_t$ for loss function:\\
	\begin{equation*}
	\min\limits_{G_t}\mathbb{E}_{y\sim s_{t-1}(y)}\mathbb{E}_{u\sim unif(0,1)}[ \|G_{t}(u,y)-G_{t-1}(u,y)\|^2]
	\end{equation*}
	given $G_{t-1}$ is  $G_t(u,y)=G_{t-1}$ for all $u$ and $y \in \Omega(s_{t-1}(y))$.
\end{lemma}
\noindent\textbf{Proof}:\\
\begin{equation*}
\begin{aligned}
&\min\limits_{G_t}\mathbb{E}_{y\sim s_{t-1}(y)}\mathbb{E}_{u\sim unif(0,1)}[ \|G_{t}(u,y)-G_{t-1}(u,y)\|^2]\\
=&\min\limits_{G_t}\int_y s_{t-1}(y)\int_u \|G_t(u,y)-G_{t-1}(u,y)\|^2 dudy\\
\geq&\int_y s_{t-1}(y)\int_u \min\limits_{G_t}\|G_t(u,y)-G_{t-1}(u,y)\|^2 dudy 
\end{aligned}
\end{equation*}
When $G_t(u,y)=G_{t-1}(u,y)$ for all $u,y$ the inequality becomes equality. \qed

\noindent
\begin{lemma} [\textbf{Digesting Loss Learns correct distribution}]\label{optgan}
	Suppose discriminator $D^{k}_t, k\in [{K_t}]$ always behave optimally and let $q_t(x|y)$ be the distribution of $G_t(u,y)$, the  the optimal $G_t(u,y)$ for digesting loss:
	\begin{equation*}
	\begin{aligned}
	\min\limits_{G_t}&\max\limits_{D^{1:{K_t}}_t} \sum_{k=1}^{{K_t}} \pi^k_t\mathbb{E}_{y\sim g_t^k(y)} \{\mathbb{E}_{x\sim p(x|y)}[ \log D^k_t(x,y)]\\
	&+ \mathbb{E}_{u\sim unif(0,1)}[\log(1-D^k_t(G_{t}(u,y),y))]\}
	\end{aligned}
	\end{equation*}
	is $q_t(x|y)=p(x|y)$ for all $y \in \Omega(g_t(y))$.
\end{lemma}
\noindent\textbf{Proof}:\\

Similar to \cite{goodfellow2014generative}, we first analyze the behavior of optimal discriminators w.r.t a fixed generator.
\begin{equation*}
\begin{aligned}
&\max\limits_{D^{1:{K_t}}_t}Loss(D_t)=\max\limits_{D^{1:{K_t}}_t}\sum_{k=1}^{{K_t}}\pi^k_t\int\limits_{y} g^k_t(y)\int\limits_{x} p(x|y)log D^k_t(y,x)\\
&\;\;\;\;\;\;\;\;\;\;\;+q_t(y|x)log(1-D^k_t(y,x)) dxdy\\
&\leq \sum_{k=1}^{{K_t}}\pi^k_t\int\limits_{y} g^k_t(y)\int\limits_{x} \max\limits_{D_t} \{p(x|y)log D^k_t(x,y)+q(y|x)log(1-D^k_t(x,y)) \}dxdy
\end{aligned}
\end{equation*}
by setting $D^k_t(y,x)=\frac{p(x|y)}{p(x|y)+q_t(x|y)}$ for all $y\in \Omega(g_t^k(y))$ we can make the inequality hold with equality.  Given a consistent optimal discriminator in each step of optimization process, the loss function of generator becomes:\\
\begin{equation*}
\begin{aligned}
Loss(G_t)= &\sum_{k=1}^{{K_t}}\pi^k_t \mathbb{E}_{y\sim g^k_t(y)}\{\mathbb{E}_{x\sim p_(x|y) [logD^k_t(x,y)]}+\mathbb{E}_{\hat{x}\sim q(x|y)} [log(1-D^k_t(x,y))]\}\\
&\Longleftrightarrow\\
Loss(q_t,\gamma)=& \sum_{k=1}^{{K_t}}\pi^k_t\int\limits_{y} g_t(y) \int\limits_{x} p(x|y)log\frac{p(x|y)}{p(x|y)+q_t(x|y)}\\
&+q_t(x|y)log\frac{q_t(x|y)}{p(x|y)+q_t(x|y)} dx +\int_{x} q_t(x|y)-1 dx dy\\
\end{aligned}
\end{equation*}
where $\gamma$ is  Lagrangian Multiplier for constraint $\int_{x} q_t(x|y) dx =1$. We have:\\
\begin{equation*}
\begin{aligned}
Loss(q_t,\gamma) \belowgeq\limits_{*} & \int\limits_{y} g_t(y) \int\limits_{x} \min\limits_{q_t} p(x|y)log\frac{p(x|y)}{p(x|y)+q_t(x|y)}\\
&+q_t(x|y)log\frac{q_t(x|y)}{p(x|y)+q_t(x|y)} + \gamma q_t(x|y)-\gamma \; dxdy
\end{aligned}
\end{equation*}
Minimizing $p(x|y)log\frac{p(x|y)}{p(x|y)+q_t(x|y)}+q_t(x|y)log\frac{q_t(x|y)}{p(x|y)+q_t(x|y)} + \gamma q_t(x|y)$ requires $\frac{p(x|y)}{p(x|y)+q_t(x|y)}$ to be constant for all possible value of $x$ and $y$. Such constraint enforces $p(x|y)=q_t(x|y)$ and $\gamma=-\log2$, which makes inequality  $*$ holds with equality. \qed
Above two lemmas describes the behavior of digesting loss and reminding loss separately. In next theorem, we show that the design of loss can work cooperatively when mixtured thus the overall loss function leads to a correct global distribution.
\begin{theorem}
	Suppose the generator has enough model capacity to obtain $q_1(x|y)=p(x|y)$ for all  $y \in g_1(y)$ and the loss $V_\tau(G_\tau,D_\tau)$ defined in Equation \ref{loss_main} is optimized optimally for each $\tau \in [t]$, then $q_t(x|y) =p(x|y)$ for all $y\in \Omega_t$.
\end{theorem}
\textbf{Proof}:\\
We will rely on induction for proof of the statement. The statement is true for $t=1$ according to our assumption and the fact that $g_1(y)=s_1(y)$. Assuming $q_{t-1}(x|y)=p(x|y)$ for all $y\in \Omega_{t-1}$, we will show $q_{t}(x|y)=p(x|y)$ for all $y\in \Omega_{t}$. Formally:
\begin{equation*}
\begin{aligned}
&V_t(G_t,D_t)=\min\limits_{G_t}\max\limits_{D^{1:{K_t}}_t}\mathbb{E}_{y\sim g_t(y)} \{\mathbb{E}_{x\sim p(x|y)}[ \log D^k_t(x,y)]\\
&+ \mathbb{E}_{u\sim unif(0,1))}[\log(1-D^k_t(G_{t}(u,y),y))]\}\\
&+ \lambda \min\limits_{G_t}\mathbb{E}_{y\sim s_{t-1}(y)}\mathbb{E}_{u\sim unif(0,1)}[ \|G_{t}(u,y)-G_{t-1}(u,y)\|^2]\\
&=\min\limits_{q_t} \int\limits_{y \in y\in \Omega(g_t(y))}\sum_{k=1}^{{K_t}}\pi^k_t g^k_t(y) \int\limits_{x} p(x|y)log\frac{p(x|y)}{p(x|y)+q_t(x|y)}\\
&+q_t(x|y)log\frac{q_t(x|y)}{p(x|y)+q_t(x|y)}  dx dy\\
&+\min\limits_{G_t}\lambda\int\limits_{y\in \Omega_{t-1}} s_{t-1}(y)\int_u \|G_t(u,y)-G_{t-1}(u,y)\|^2 dudy\\
&\belowgeq\limits_{*} \int\limits_{y \in y\in \Omega(g_t(y))}\sum_{k=1}^{{K_t}}\pi^k_t g^k_t(y) \int\limits_{x} \min\limits_{q_t} p(x|y)log\frac{p(x|y)}{p(x|y)+q_t(x|y)}\\
&+q_t(x|y)log\frac{q_t(x|y)}{p(x|y)+q_t(x|y)}  dx dy\\
&+\lambda\int\limits_{y\in \Omega_{t-1}} s_{t-1}(y)\int_u \min\limits_{G_t}\|G_t(u,y)-G_{t-1}(u,y)\|^2 dudy\\
\end{aligned}
\end{equation*}
Next we show the inequality $*$ attains equality if $q_t(x|y)=p(x|y)$ for all $y \in \Omega_t$. First we note that for $y \in \Omega(g_t(y)) \cap \Omega_{t-1}$  the digesting loss and reminding loss shares the same optimal solution. Note $G_t(u,y)=G_{t-1}(u,y)$ is equivalent to $q_t(x|y)=q_{t-1}(x|y)$ since $G_t$ and $G_{t-1}$ shares the same random seed $u$. We have for $y \in \Omega(g_t(y)) \cap \Omega_{t-1}$, $q_t(x|y)=q_{t-1}(x|y)=p(x|y)$ due to the inductive assumption for reminding loss. The optimality of $q_t(x|y)=p(x|y)$ is due to Lemma \ref{optgan}.

Next we have $q_t(x|y)=p(x|y)$ for $y \in \Delta \Omega_t$. For $y \in \Omega(g_t(y))-\Omega(s_{t-1}(y))$, $q_t(x|y)=p(x|y)$ according to Lemma \ref{optgan}. 
For $y \in \Omega_{t-1}-\Omega(g_t(y))$, $G_t(u,y)=G_{t-1}(u,y)$ according to Lemma \ref{optrmd}. \qed

\end{document}